\RequirePackage{fix-cm}
\documentclass[smallextended]{svjour3}       
\usepackage{graphicx}
\usepackage[numbers,round]{natbib}
\journalname{J Ambient Intell Human Comput}

\makeatletter
\renewcommand\@biblabel[1]{}
\renewcommand\NAT@bibsetnum[1]{\settowidth\labelwidth{\@biblabel{#1}}%
   \setlength{\leftmargin}{\bibindent}\addtolength{\leftmargin}{\dimexpr\labelwidth+\labelsep\relax}%
   \setlength{\itemindent}{-\bibindent}%
   \setlength{\listparindent}{\itemindent}
\setlength{\itemsep}{\bibsep}\setlength{\parsep}{\z@}%
   \ifNAT@openbib
     \addtolength{\leftmargin}{\bibindent}%
     \setlength{\itemindent}{-\bibindent}%
     \setlength{\listparindent}{\itemindent}%
     \setlength{\parsep}{0pt}%
   \fi
}
\makeatother

\begin{document}

\title{Prediction of Seasonal Temperature Using Soft Computing Techniques: Application in Benevento (Southern Italy) Area\thanks{In collaboration with the Environmental Data Processing Course Group of Universit\`a del Sannio: Fabio Caporaso, Antonio Di Mezza, Guido Razzano, Pasquale Clemente, and Alessio Giardiello.}
}

\titlerunning{Prediction of Seasonal Temperature Using Soft Computing Techniques}        

\author{Salvatore Rampone         \and
       Alessio Valente 
}


\institute{\at
              Department of Science and Technology (DST) \\
              Universit\`a del Sannio\\
              Via dei Mulini 59/A, 
Benevento, Italy, I-82100\\
              \email{\{rampone, valente\}@unisannio.it}           
}

\date{Received: date / Accepted: date}

\maketitle
\begin{abstract}
In this work two soft computing methods, Artificial Neural Networks and Genetic Programming, are proposed in order to forecast the mean air temperature that will occur in future seasons. The area in which the soft computing techniques were applied is that of the surroundings of the town of Benevento, in the south of Italy, having the geographic coordinates (lat. 41$^{\circ}$07'50''N; long.14$^{\circ}$47'13''E). This area  is not affected by maritime influences as well as by winds coming from the west. The methods are fed by data recorded in the meteorological stations of Benevento and Castelvenere, located in the hilly area, which characterizes the territory surrounding this city, at 144 m a.s.l. Both the applied methods show low error rates, while the Genetic Programming offers an explicit rule representation (a formula) explaining the prevision.

\end{abstract}

\section{Introduction}

Air temperature is a measure of how hot or cold the air is. The knowledge of the mean air temperature of the future seasons is very helpful for individuals, but even more for organizations whose workers and machines have to operate outdoors (agriculture, armed force, railways, maintenance of roads, tourism, etc.) and for gas and electricity companies. 
The seasonal temperature values, in fact, have a great economic impact for both organizations and companies, and their prediction would allow a more efficient activity planning.


The accurate prediction of the air temperature is very complex and usually requires a processing of a large amount of data on the weather conditions as well as the scientific understanding of atmospheric processes and their possible spatial and temporal variations~\cite{Sardaetal1996,Andersonetal1999,Stockdale2000}. Moreover, it is not possible to perform the prediction of the seasonal temperature with a minimum error for all the months and places in the world. Generally, the quality of seasonal forecasting has reached a good reliability over the most of the tropical regions, while the extra tropical latitudes have achieved reliable results only in certain regions (for example, in Northern America). More specifically, in Europe the predictability of temperature is quite complicated due to the variability and external influences of the local climate~\cite{RodwellandDoblas-Reyes2006}. However in the recent decades a considerable effort has been made to improve the understanding of physical phenomena responsible for the seasonal variability of climate parameters~\cite{Sahaetal2006,InesonandScaife2009}.

In order to forecast the temperature that will occur in future seasons, climatological models are frequently used. Some of them follow an empirical approach based on several series of historical data, useful for building predictive models; others develop a theoretical approach on complex calculations with parameters and functions that are able to understand the behaviour of the atmospheric system. 

The first approach considers a set of minimum, maximum and mean temperatures recorded for several decades (at least thirty years) in representative meteorological stations in the area, while the second approach uses numerical analysis techniques which involve data observed by radiosondes, weather satellites and surface weather observations of land and oceans processed by powerful supercomputers, which are necessary because of the huge amount of data. For the latter approach, the horizontal spatial domain is global, if it covers the entire Earth, or regional, if it covers only a small part of the planet. In the first approach, the forecast is in somewhat approximate, although it is indicative of the future trend of the seasonal temperature in a certain area; in the second approach the values of minimum and maximum temperature of the foreseen seasons can be estimated very reliably, because it takes into account the correlations of anomalies in the atmospheric circulation at large scale. However, it is difficult for operators to arrange the  numerical approach on a smaller scale for planning purposes.

In order to make the forecasting of foreseen seasonal temperatures less sophisticated, in this work two soft computing methods are applied: Artificial Neural Networks (ANNs)~\cite{Bishop1996,Haykin2008} and Genetic Programming (GP)~\cite{Koza1992}. 

As ANN we use a feedforward Multi Layer Perceptron (MLP) Neural Network, a computational structure made by many processing elements (units), the neurons, operating in parallel, aiming to approximate an unknown function of its inputs~\cite{BealeandJackson1990}. 
To configure the MLP we use the training procedure called ``back propagation''~\cite{BealeandJackson1990}. It uses a subset of the data set feature vectors, each one labeled with the correct output, as examples of the correct input/output relationship. Artificial Neural Networks have already been successfully applied in different contexts~\cite{Rampone2013, Ramponeetal2013, RamponeandValente2012}, including the atmospheric researches, such as the prediction of maximum daily precipitation~\cite{Nastosetal2014}, the monthly precipitation and evapotranspiration index~\cite{DeoSahin2015} and air temperature~\cite{Sahin2012,Venkadeshetal2013}. Generally, in these works, in addition to the change of the reference framework, which can be on a local or regional scale, a wide range of parameters is used. The use of parameters depends on the availability of data. It is not always possible to have them at the local level as well as the reduction in a local scale of the information available at the regional scale cannot be free from errors and approximations. 

We also employ GP in order to improve the understanding of the neural method performance. GP is an evolutionary computational technique proposed by Koza~\cite{Koza1992,Koza1994} in order to extract automatically intelligible relationships in a system without being explicitly programmed. It has been used in many applications such as symbolic regressions~\cite{Davidsonetal2003}, and classifications~\cite{DeStefanoetal2002,ZhangandBhattacharyya2004}. It works by using genetic algorithms~\cite{Goldberg1989} to generate and evolve automatically composite functions, traditionally represented as tree structures~\cite{Cramer1985}. GP is then able to show the relation between input data and output data by means of an explicit formula. Also this soft computing method has been successfully applied to many problems of practical order~\cite{Makkeasornetal2008, Shirietal2012, Stanislawskaetal2012, g2007}. 

This paper is organized in the following way: in the next Section we describe the application area and in Section 3 the environmental data used for tuning the methods. The results are reported in the Experiments Section.

\section{Application Area}
The area in which soft computing techniques are applied is that of the surroundings of the town of Benevento having the geographic coordinates (lat. 41$^{\circ}$ 07' 50'' N; long. 14$^{\circ}$ 47' 13'' E). It is located equidistant from the seas and protected by the hills located in the west that is an inner area of the Italian peninsula. Therefore, the area of this application is not affected by maritime influences, nor by winds coming from the west. 
Nevertheless, its geographic location falls in the context of a Mediterranean climate with a tendency to continental features.
More specifically, along the Italian peninsula it is observed, in winter, the interaction between the area flows from the northwest and west and those who move from the east, which causes heavy rains, while in summer the presence of subtropical anticyclone determines a prolonged drought interrupted by sporadic thunder phenomena. The north-western winter air streams can bring down the minimum temperature to a few degrees above zero, and the maximum in summer exceeds 30$^{\circ}$C.

This synoptic situation is made evident by the data recorded in the meteorological station of Benevento~\cite{CAR2016}. It is located in the hilly area, which characterizes the territory surrounding this town, at 144 m a.s.l. Although it has recorded rainfall and temperature data since 1882, the official ones published on the Hydrological Annals are available only from 1925 until 1998 with occasional missing data. 
The decommissioning of the Benevento station forced us to examine the data of another site, at a distance of less than twenty kilometers north-west from Benevento. So the data corresponding to the last 13 years come from the Castelvenere station (lat. 41$^{\circ}$14'15''N; long. 14$^{\circ}$33'16''E), that is located in a similar geographical position (140 m a.s.l.) and it presents a minimal difference with the data of temperature and precipitation recorded in Benevento. However, in order to avoid mistakes, these data have been scaled according with this difference. In both stations, for each year the available data are the daily minimum temperature, the daily maximum temperature, the daily mean temperature, the daily amount of rainfall and the number of rainy days that occurred during each month.

In order to better show the weather characteristics of the area, we provide some information obtained from the considered series. The mean maximum temperature recorded during the summer season varies from 25.6$^{\circ}$C to 32.1$^{\circ}$C. The mean minimum and maximum value has been reached respectively in 1976 and 1928. The mean minimum temperature in the winter season ranges from a low of -0.8$^{\circ}$C in 1982 to a maximum of 7.8$^{\circ}$C in 1979. Also in 1982, the highest annual temperature range of 31.1$^{\circ}$C was reached, although the yearly mean temperature remains below 17$^{\circ}$C. The daily mean temperature is slightly below 15$^{\circ}$C. 

Just to complete the data relating to the maximum temperatures recorded in the last decades, Castelvenere station shows a tendency to increase with peaks that have exceeded abundantly 40$^{\circ}$ during summer months, whereas in the same station the minimum temperature has gone down to 7$^{\circ}$ below zero during the winter months. The mean maximum temperature of the summer months is within the same range defined before, but basically the values are higher. Even the mean minimum temperature of the winter months falls between the extremes defined at a general level, however, it should be noticed that it is not unusual to have periods of more than one week with temperatures below zero (2009 and 2012).
As for precipitations, it can be seen clearly in the annual distribution of the data that they are highest in autumn and winter seasons and at a minimum in the summer. In particular, the range of the annual precipitation varies from a minimum of 256 mm in 2003 to a maximum of 1127 mm in 1955, with an average value which is near to 800 mm. The largest quantities were mainly recorded in the autumn months (nearly 600 mm in 2010 and over 500 mm in 1966), while the lowest ones were recorded in the summer months (20 mm in 2000 and 21 mm in 1987). This effect is confirmed by the frequency of rainfall, which sees a concentration of rainy days in the autumn months (56 rainy days in 2002), in contrast to summer months in which it is reduced dramatically, as in the months of the beginning of this century with one day in 2000.

\section{Data}
In the proposed application we want to predict the mean seasonal temperature of the subsequent years.  Therefore, a number of parameters (features) to assess the seasonal conditions of this area are calculated from the available data from 1925 to 2012. Such parameters are then used for the prediction. According to the data availability in the series of the meteorological stations in Benevento area, we consider 9 features: 

\begin{table}[ht]
\caption{Parameters calculated from the available data from 1925 to 2012.}

{\begin{tabular}{ll}
\hline\noalign{\smallskip}
Feature & Range \\
\noalign{\smallskip}\hline\noalign{\smallskip}
Year $(Y)$	&From 1925 to 2012\\
Season $(S)$	&Winter, Spring, \\
&Summer, Autumn\\
Mean Seasonal Temperature $(MST)$	&C degrees\\
Mean Seasonal of Daily Maximum Temperatures $(MSDMT)$	&C degrees\\
Mean Seasonal of Daily minimum Temperatures $(MSDmT)$	&C degrees\\
Mean Yearly Temperature $(MYT)$	&C degrees\\
Mean Seasonal Rainfall $(MSR)$	&mm\\
Number of Seasonal Raining Days $(NSRD)$	&Positive Integer\\
Mean Seasonal Temperature in the Next Year $(MSTNY)$ 	&C degrees \\
\noalign{\smallskip}\hline

\end{tabular} 
\label{ta1}}
\end{table}

With the exception of the Season $(S)$ all the features are expressed as numeric variables. They should contribute to the prediction of seasonal mean temperature of the year following the year in progress. In addition to temperature values, which express the most features used, two sets of data concerning the precipitation are considered. They have been inserted in the input data, since the temperature is to be put in relation to the general circulation and disturbances that could be incurred. In some cases the occurrence of rainfall can be enforced by local situations, such as mountain range and vegetation, which determine the amount of moisture in the air. Such amount is to be put in relation with the temperature that is observed. Furthermore, rainfall is also related to the number of rainy days, useful to detect the frequency with which this phenomenon occurs. 

Missing data are restricted to the period of the Second World War (1942 1948) and the failure of the stations for certain periods; however, this has not compromised the quality of information used for the application. To obviate these data, it has not been made any interpolation, while for the data of the second station an appropriate correction has been introduced taking into account the local influences on the recorded data (different exposure to the Solar Radiation, moisture conditions, etc.).

The resulting dataset is composed of 248 labeled examples, where each example has the following structure
\begin{equation}
Xp; tp = (Y; S; MST; MSDMT; MSDmT; MYT; MSR; NSRD); MSTNY 
\end{equation}

\section{Experiments}
\subsection{MLP}
According to the 10-fold cross-validation methodology \cite{DevijverandKittler1982}, the whole data set has been partitioned into 10 non overlapping subsets. Of these 10 subsets, one can be chosen to evaluate the neural network performance (validation set), and the remaining 9 can be used to instruct it (training sets), i.e. setting the network weights. This procedure has been repeated 10 times, corresponding to all possible choices of the validation set and the corresponding average performance has been gauged in terms of an average \%-misclassification error. Cross-validation is important in testing hypotheses suggested by the data, especially where further samples are costly or impossible to collect.

\begin{figure}[phtb]
\centerline{\includegraphics[width=10.7cm]{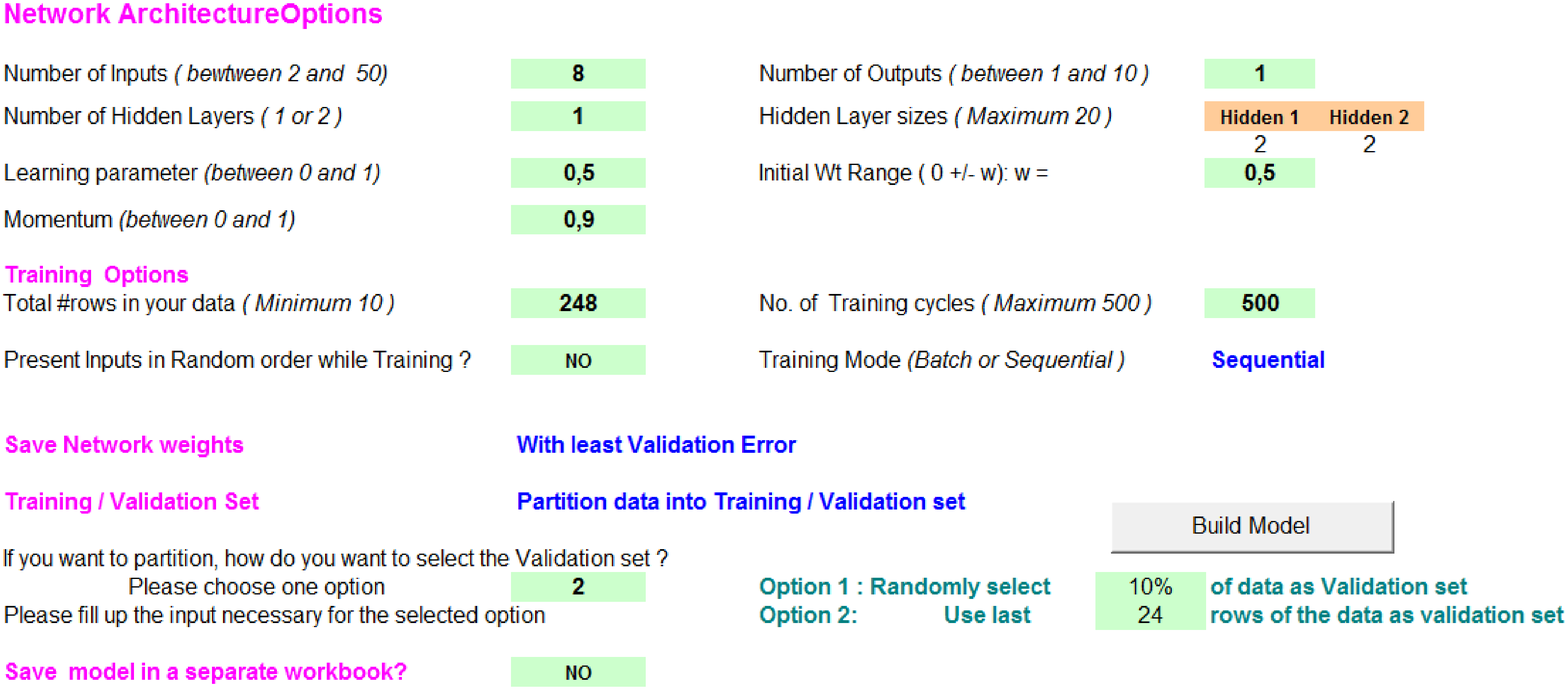}}
\vspace*{8pt}
\caption{Parameter set by using the Excel-based system which simulates the MLP neural network. \label{f4}}
\end{figure}

For our experiments we adopted an Excel-based system which simulates a neural network developed by Angshuman Saha, that is freely available all over the web\footnote{http://xoomer.virgilio.it/srampone/NNpred01.zip.}. By using this tool, we can define feed-forward back-propagation networks with 1 or 2 hidden layers, enter training data, set various learning parameters, start a learning phase and see the results (e.g. error rate) presented in various ways, for example graphically (see Fig.~\ref{f4}). Categorical values are automatically converted into numerical ones. The weights $W_{kn}$ are initially randomly chosen in a fixed range. The error back propagation algorithm has several parameters, and the most important ones are called, respectively, 
``learning rate'' and ``momentum term''. The first term is a measure of the influence degree, in the formula for updating weights, of the error term, whereas the latter one determines the influence of the past history of weight changes in the same formula.

One of the most important characteristics of a MLP is the number of neurons in the hidden layer(s). If an inadequate number of neurons are used, the network will be unable to model complex data, and the resulting fit will be poor. If too many neurons are used, the network generalizes poorly to new, unseen data. So the hidden layers have been configured by a random initial choice and
\begin{itemize}
\item	 pruning, i.e. eliminating nodes, when the network learns easily the training set but shows bad performances on the validation set; and
\item	 growing, i.e. adding nodes when the network shows bad performances on the training set.
\end{itemize}
In our case the resulting architecture is made up of 8 inputs, a single hidden layer of 2 neurons, and one output.
 
The number of training cycles (epochs) has been fixed to 500. 

To value the performances, we use the error percentage resulting from the 10-fold cross-validation methodology and the coefficient of determination, denoted $R^2$, that indicates the proportion of the variance in the dependent variable that is predictable from the independent variable.

\begin{figure}[phtb]
\centerline{\includegraphics[width=7.7cm]{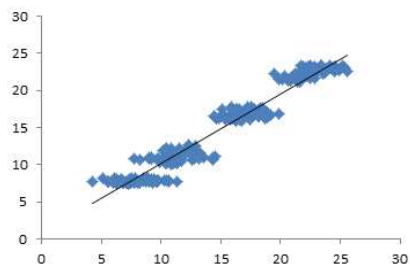}}
\vspace*{8pt}
\caption{Plot of Expected (X-axis) vs Predicted (Y-axis) temperatures on the whole data set by using MLP. \label{f5}}
\end{figure}

The results on the whole data set have been reported in Table 2. The mean error is of 8,71\% while the coefficient of determination is 0,97, a very high value.  Fig.~\ref{f5} reports comparatively the predicted and expected temperature values. 

\begin{table}[ht]
\caption{Ten fold cross validation by using the selected architecture on the whole data set}
\begin{tabular}{@{}ccc@{}} 
\hline\noalign{\smallskip}
Trial	&Validation error	&$R^2$\\
\noalign{\smallskip}\hline\noalign{\smallskip}

1	&3,68\%	&0,97\\
2	&8,31\%	&0,97\\
3	&7,70\%	&0,97\\
4	&9,28\%	&0,97\\
5	&7,32\%	&0,97\\
6	&6,15\%	&0,97\\
7	&16,41\%	&0,97\\
8	&10,35\%	&0,97\\
9	&10,41\%	&0,97\\
10	&7,51\%	&0,97\\
\noalign{\smallskip}\hline\noalign{\smallskip}
{\bf Mean}	&8,71\%	&0,97\\

\noalign{\smallskip}\hline

\end{tabular} \label{ta2}
\end{table}

Then, since the data originated from two different stations, we partitioned the dataset, according to its origin, in two subsets, i.e. from 1925 al 1998 and from 2000 to 2012, and the ten fold cross validation procedure has been repeated for each one. The results on the two subsets have been reported respectively in Table 3 and Table 4.
\begin{table}[ht]
\caption{Ten fold cross validation by using the selected architecture on the subset 1925-1988.}
\begin{tabular}{@{}ccc@{}} 
\hline\noalign{\smallskip}

Trial	&Validation error	&$R^2$\\
\noalign{\smallskip}\hline\noalign{\smallskip}
1	&10,76\%	&0,97\\
2	&7,73\%	&0,97\\
3	&8,85\%	&0,97\\
4	&6,79\%	&0,97\\
5	&8,21\%	&0,97\\
6	&8,63\%	&0,96\\
7	&7,10\%	&0,97\\
8	&12,52\%	&0,97\\
9	&11,87\%	&0,97\\
10	&13,82\%	&0,97\\
\noalign{\smallskip}\hline\noalign{\smallskip}
{\bf Mean}	&9,63	\%	&0,97\\
\noalign{\smallskip}\hline
\end{tabular} \label{ta3}
\end{table}

\begin{table}[ht]
\caption{Ten fold cross validation by using the selected architecture on the subset 2000-2012.}
\begin{tabular}{@{}ccc@{}} 
\hline\noalign{\smallskip}

Trial	&Validation error	&$R^2$\\
\noalign{\smallskip}\hline\noalign{\smallskip}
1	&4,19\%	&0,97\\
2	&5,89\%	&0,97\\
3	&8,78\%	&0,97\\
4	&7,22\%	&0,97\\
5	&7,48\%	&0,97\\
6	&3,29\%	&0,97\\
7	&3,21\%	&0,97\\
8	&5,03\%	&0,97\\
9	&1,04\%	&0,97\\
10	&3,79\%	&0,97\\
\noalign{\smallskip}\hline\noalign{\smallskip}
{\bf Mean}	&4,99\%	&0,97\\
\noalign{\smallskip}\hline
\end{tabular} \label{ta4}
\end{table}

\subsection{GP}

Given the data set characteristics, we are looking for a formula $f()$ that satisfies
\begin{equation}
MSTNY = f(Y; S; MST; MSDMT; MSDmT; MYT; MSR; NSRD)
\end{equation}
We limit the set of component functions to the arithmetic operators $(+, -, *, /)$ plus some trigonometric functions (sine, cosine and tangent), the exponential and the natural logarithm, the logistic function, the hyperbolic tangent and the minimum function.

As fitness measure we use the mean square error $(MSE)$.

The exemplars are split in two groups for training (90\%) and validation (10\%). The training set is created by randomly selecting samples from the whole dataset.

We use a genetic programming software tool called Eureqa, for detecting equations and hidden mathematical relationships in a given data set. Eureqa works in order to reduce the error function given by the discrepancy between the data and the generated model~\cite{SchmidtandLipson2009}. Eureqa is a proprietary A.I.-powered modeling engine originally created by Cornell's Artificial Intelligence Lab and later commercialized by Nutonian, Inc. The software uses evolutionary search to determine mathematical equations that describe sets of data in their simplest form. It also allows to evidence the behaviour of each solution with respect to its size.

\begin{figure}[phtb]
\centerline{\includegraphics[width=7.7cm]{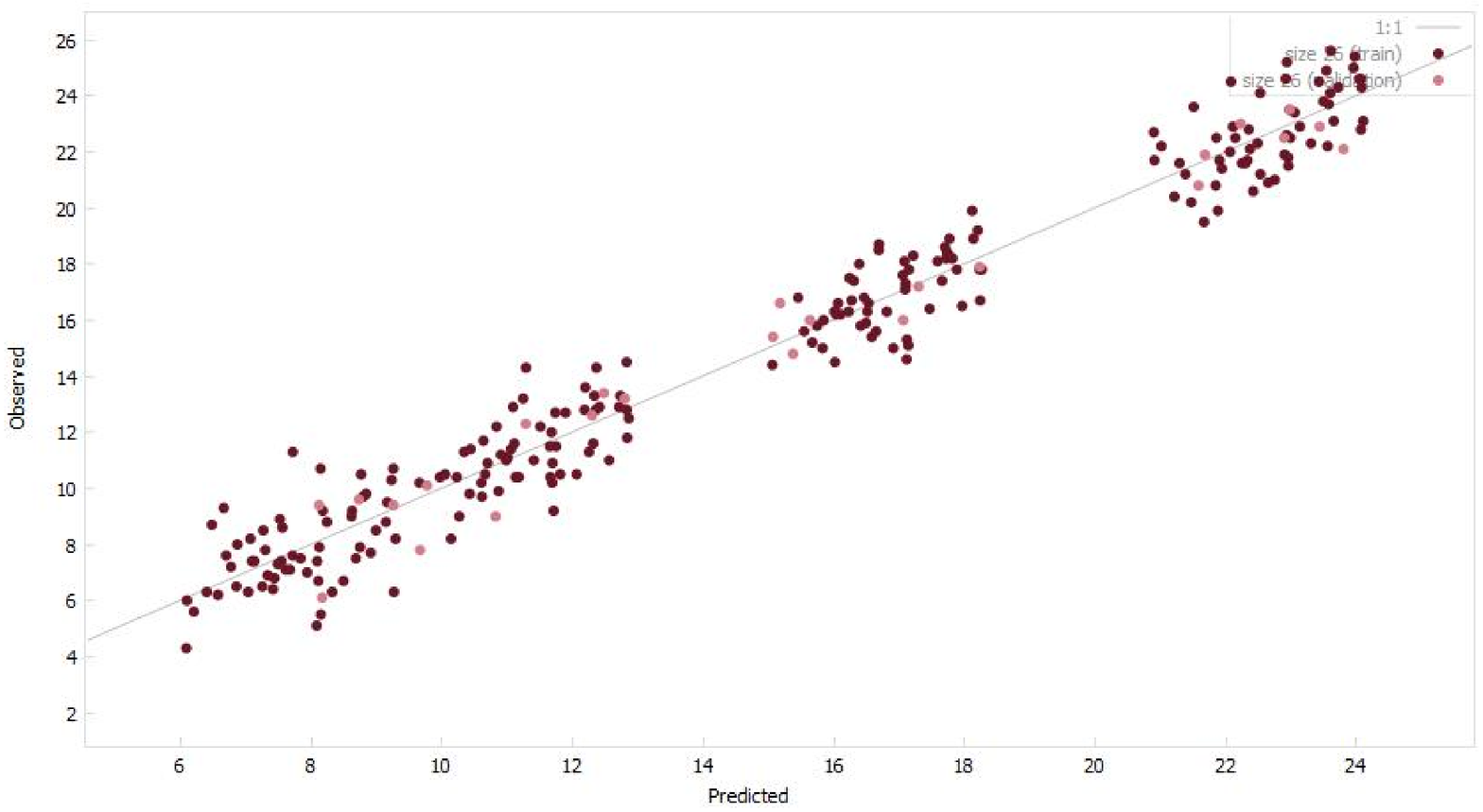}}
\vspace*{8pt}
\caption{Plot of Expected (X-axis) vs Predicted (Y-axis) temperatures on the whole data set by using GP. \label{f6}}
\end{figure}

After about 700000 generations we found a set of solution, as reported in Table 5. The performance of each solution are reported in Table 6. A plot of Expected vs Predicted temperatures on the whole data set by using solution 10 is reported in Fig.~\ref{f6}.

\begin{table}[ht]
\caption{Ten fold cross validation by using the selected architecture on the subset 2000-2012.}
{\begin{tabular}{@{}cccl@{}} 
\hline\noalign{\smallskip}

Solution num.	&Size	&Fit	&Solution\\
\noalign{\smallskip}\hline\noalign{\smallskip}
1	&1	&1,000	&$MSTNY = MST$\\
2	&3	&0,985	&$MSTNY = 0,99*MST$\\
3	&6	&0,806	&$MSTNY = MST - cos(MSDMT)$\\
4	&5	&0,982	&$MSTNY = 0,51 + 0,96*MST$\\
5	&10	&0,762	&$MSTNY = 0,55 + 0,95*MST - cos(MSDMT)$\\
6	&14	&0,445	&$MSTNY = MYT - 6,67 - 5,28*S*cos(0,55 + S)$\\
7	&38	&0,309	&$MSTNY = MYT - 7,00 - cos(4884,11*Y)*$\\
&&& $min(cos(0,48*Y), cos(0,28*Y)) - 5,41*S*cos(0,54 + S)$\\
8	&32	&0,340	&$MSTNY = 0,84*MYT - 4,34 - cos(4884,11*Y)*$\\ 
&&&$cos(2,584 + 0,84*MYT)- 5,32*S*cos(0,54 + S)$\\
9	&28	&0,350	&$MSTNY = MYT - 6,66 - cos(4884,10*Y)*$\\
&&&$cos(0,58*MYT) - 5,32*S*cos(0,54 + S)$\\
10	&26	&0,436	&$MSTNY = 7,77 + 0,64*cos(4951,06*Y) $\\
&&&$- cos(MYT) - 5,33*S*cos(0,54 + S)$\\
\noalign{\smallskip}\hline
\end{tabular} \label{ta5}}
\end{table}

\begin{table}[ht]
\caption{Performance of the solution set in terms of $R^2$ and $MSE$.}
{\begin{tabular}{@{}ccccccccccc@{}} 
\hline\noalign{\smallskip}

Solution&	1	&2	&3	&4	&5	&6	&7	                &8	&9	&10\\
\noalign{\smallskip}\hline\noalign{\smallskip}
{\bf $R^2$}&0,92	&0,92	&0,92	&0,94	&0,94	&0,96	&0,97	&0,97	&0,97	&0,98\\
{\bf $MSE$}	&2$\times10^4$	&2$\times10^7$	&2$\times10^7$	&2$\times10^7$	&2$\times10^6$	&1,00	&0,98	&0,78	&0,76	&0,69\\
\noalign{\smallskip}\hline
\end{tabular} \label{ta6}}
\end{table}

As it was evidenced by the solution expressions, just 3 parameters determine the solution. We value the relevance of each variable in terms of

{\bf Sensitivity}: The relative impact within this model that a variable has on the target variable.

{\bf \% Positive}: The likelihood that increasing this variable will increase the target variable. 

{\bf Positive Magnitude}: When increases in this variable lead to increases in the target variable, this is generally how big the positive impact is.

{\bf \% Negative}: The likelihood that increasing this variable will decrease the target variable

{\bf Negative Magnitude}: When increases in this variable lead to decreases in the target variable, this is generally how big the negative impact is.

The results, for the best solution found are reported in Table 7.

\begin{table}[ht]
\caption{Variable Sensitivity in the solution 10.}
{\begin{tabular}{@{}cccccc@{}} 
\hline\noalign{\smallskip}

Variable&	Sensitivity	&\% Positive	&Positive Mag.&	\% Negative	&Negative Mag.\\
\noalign{\smallskip}\hline\noalign{\smallskip}
$Y$	&3,0141	&46\%	&3,3674	&54\%	&2,7143\\
$S$	&1,3642	&67\%	&1,1636	&33\%	&1,7654\\
$MYT$	&0,1883	&100\%	&0,1883	&0\%	&0\\
\noalign{\smallskip}\hline
\end{tabular} \label{ta7}}
\end{table}

Then the experiment has been repeated by adding the Gaussian function to the set of functions. In this case the best solution considers more and different variables, as reported in Table 8.

\begin{table}[ht]
\caption{Variable Sensitivity in the solution MSTNY = 0,6354* MYT - 0,183 - 0,07226*tan(MYT) - 0,0005114* MYT * MSDMT S2 - 0,901* MST *sin(2,203 + 0,9123*S),}
{\begin{tabular}{@{}cccccc@{}} 
\hline\noalign{\smallskip}

Variable&	Sensitivity	&\% Positive	&Positive Mag.&	\% Negative	&Negative Mag.\\
\noalign{\smallskip}\hline\noalign{\smallskip}
$S$	&1,1779	&67\%	&1,1357	&33\%	&1,2622\\
$MST$	&0,55971	&100\%	&0,55971	&0\%	&0\\
$MSDMT$&	0,41929	&0\%	&0	&100\%	&0,41929\\
$MYT$	&0,16345	&80\%	&0,12137	&20\%	&0,33176\\
\noalign{\smallskip}\hline
\end{tabular} \label{ta8}}
\end{table}

\section{Discussion}
In this study, two soft computing methods, as Artificial Neural Networks, in the Multi-Layer Perceptron $(MLP)$ approach, and Genetic Programming $(GP)$, were applied to predict next year's seasonal temperatures. In both methods we considered the same weather variables extracted from the records of two meteorological stations of Benevento and Castelvenere. While the traditional statistical methods need several a priori assumptions, $MLP$ and $GP$ are non-parametric tools and suitable for any situations and data sets.

Both the applied methods show low error rates and high correlation values.

As regard the application of $MLP$, the results on the whole data set show a mean error of 8,71\% with a very high value of coefficient of determination of 0,97. If we consider only the subset 2000-2012, the ten fold cross validation result is lower than 5\% with the same coefficient of determination. 

However, from a customer manager’s point of view, the classification results from the $MLP$ are sub-symbolic and are often difficult to comprehend. The proposed $GP$ approach offers an explicit rule representation (a formula) while maintaining the low error rates and high correlation values.

As it has been evidenced in the last solution reported in Table 5, a formula which has a relatively low size and a rather good fitness considers just the mean yearly temperature $(MYT)$ of the previous year, the actual year $(Y)$ and the relative season $(S)$. This solution reaches a very good performance in terms of $R^2$ and $MSE$  (the former is 0,98\% and the latter is 0,69, as reported in Table 6). The comprehension of the GP results has markedly improved in the next serial of tests and it  has pointed out that the foreseen of seasonal temperature $(MSTNY)$ is related to the tendency of previous mean season temperature $(MST)$ and subordinately to the mean yearly temperature $(MYT)$ (see Table 8). This result could be linked to the atmosphere circulation acting in the application area, otherwise these variables would have a higher negative sensitivity~\cite{Stanislawskaetal2012}. A major impact of the mean seasonal of daily maximum temperatures $(MSDMT)$ and the mean seasonal of daily minimum temperatures $(MSDmT)$, as also the mean seasonal rainfall $(MSR)$ and number of seasonal raining days $(NSRD)$ in the solution test, would be evidence of anomalies in the atmosphere circulation (i.e. higher content in greenhouse gas, frequent cloud burst). 
In our application, developed in the area of Benevento, the evolutionary trend appears almost linear, so evidencing the  dominance of the natural forces. Probably this was to be expected for the natural and mainly agricultural land use, affecting to a lesser degree on the temperature behaviour with respect to industrial areas.

These good results encourage the improving of these methods, by exploring and fine-tuning the various parameters, whose exact determination may be quite complex (e.g. temperature, pressure, moist and so on). In the last decade these estimations have been further complicated by anthropogenic forces that make less comprehensible the behaviour of the temperature~\cite{Pasinietal2006}.


\begin{acknowledgements}
The authors are grateful to L. Rampone for the careful reading of the paper.
\end{acknowledgements}



\end{document}